\documentclass{article}

\usepackage[preprint,nonatbib]{neurips_2025}

\usepackage{booktabs}
\usepackage[utf8]{inputenc} 
\usepackage{hyperref}       
\usepackage{url}            
\usepackage{booktabs}       
\usepackage{amsfonts}       
\usepackage{nicefrac}       
\usepackage{microtype}      
\usepackage{xcolor}         
\usepackage[numbers]{natbib}
\usepackage{amsmath}
\usepackage{enumerate}
\usepackage[shortlabels]{enumitem}

\usepackage{multirow}
\usepackage{algorithm}
\usepackage[noend]{algpseudocode}
\usepackage{graphicx}

\usepackage{subcaption}

\newtheorem{theorem}{Theorem}[section]

\newtheorem{lemma}[theorem]{Lemma}

\title{On Kernel-based Variational Autoencoder}

\author{%
  Tian Qin \\ 
  Math department\\
  Lehigh University\\
  Bethlehem, PA 18015 \\
  \texttt{tiq218@lehigh.edu} \\
  \And
  Wei-Min Huang \\
  Math department\\
  Lehigh University\\
  Bethlehem, PA 18015 \\
  \texttt{wh02@lehigh.edu} \\
}

\begin{document}

\maketitle

\begin{abstract}
In this paper, we bridge Variational Autoencoders (VAEs) \cite{VAE} and  kernel density estimations (KDEs) \cite{Rosenblatt,Parzen} by approximating the posterior by the expectation of kernel density estimator and deriving a new lower bound of empirical log likelihood. The flexibility of KDEs provides a new perspective of controlling the KL-divergence term in original evidence lower bound (ELBO) which enhances the flexibility of the posterior and prior pairs in VAE.  We show that  the Epanechnikov kernel gives the tightest upper bound in controlling the KL-divergence under appropriate conditions \cite{Epanechnikov,Bickel} in theory and develop a kernel-based VAE called Epanechnikov Variational Autoenocoder (EVAE). The implementation of Epanechnikov kernel in VAE is straightforward  as it lies in the ``location-scale'' family of distributions where reparametrization tricks can be applied directly. Compared with Gaussian kernel, Epanechnikov kernel has compact support which should make the generated sample less blurry. Extensive experiments on benchmark datasets such as MNIST, Fashion-MNIST, CIFAR-10 and CelebA  illustrate the superiority of Epanechnikov Variational Autoenocoder (EVAE) over vanilla VAE and other baseline models in the quality of reconstructed images, as measured by the FID score and Sharpness \cite{WAE}.
\end{abstract}

\section{Introduction}
\label{introduction}
In variational inference, an autoencoder learns to encode the original data $\mathbf{x}$ using learnable function $f^{-1}_{\phi}(\mathbf{x})$ and then to reconstruct $x$ using a decoder $g_{\theta}$. \cite{VAE} proposed a  stochastic variational inference and learning algorithm called Variational Autoencoders (VAEs) which can be further used to generate new data. According to VAE, a lower bound on the empirical likelihood, which is known as ELBO, is maximized so that the fitted model $p_{\theta}(\mathbf{x}|\mathbf{z})=\mathcal{N}(\mathbf{x};g_{\theta}(\mathbf{x}),I_{D})$ and $q_{\phi}(\mathbf{z}|\mathbf{x})=\mathcal{N}(\mathbf{x};f_{\phi}(\mathbf{x}),I_{D})$  can approximate the true conditional distributions $p(\mathbf{x}|\mathbf{z})$ and $p(\mathbf{z}|\mathbf{x})$, respectively. The isotropic Gaussian  prior and posterior distribution in vanilla VAE is mathematically convenient since the corresponding ELBO is analytic. But the main drawbacks are the lack of expressibility of latent space and the possible posterior collapse. There are two popular directions for extending VAEs to address these drawbacks.

\subsection{Approximate posterior $q_{\phi}(\mathbf{z}|\mathbf{x})$:}

To enhance the posterior expressiveness, Normalizing flow (NF) \cite{JimenezRezende2015VariationalIW} applies a sequence of invertible transformations to initial density $q_{0}(\mathbf{z})$ to achieve more expressive posteriors.  $\beta$-VAE \cite{Higgins2016betaVAELB} introduced a new parameter $\beta$ in balancing the reconstruction loss with disentangled latent representation $q_{\phi}(\mathbf{z}|\mathbf{x})$. Importance weighted Autoencoders (IWAE) \cite{Burda2015ImportanceWA} improved log-likelihood lower bound by importance weighting. It also approximated the posterior with multiple samples and enriched latent space representations. 
\cite{AVAE} proposed the Aggregate Variational Autoencoder (AVAE), which  estimates the posterior by Gaussian kernel density estimations (KDEs) to improve the quality of the learned latent space.


\subsection{Prior distribution $p(\mathbf{z})$ of latent variable $\mathbf{z}$:}

Instead of approximating posterior, we can replace Gaussian prior with other flexible distributions. For instance, \cite{Dilokthanakul2016DeepUC} used a simple Gaussian mixture prior and \cite{VampPrior} introduced a mixture prior called ``VampPrior'' which  consists of a mixture distribution  with components given by variational posteriors. The possibility of  non-parametric priors is explored in \cite{nalisnick2017stickbreaking}, which utilized a truncated stick-breaking process. \cite{Hasnat2017vonMM} and  \cite{Davidson2018HypersphericalVA} attempted to replace Gaussian prior by von Mises-Fisher(vMF) distribution to cover hyperspherical latent space. For data with heavy-tailed characteristics, \cite{tVAE} replaced Gaussian prior with heavy-tailed distributions such as the Student's t-distribution, which allows the model to capture data with higher kurtosis, leading to more robust representations. \cite{CVAE} utilized optimal transport  theory  design priors that enforce a coupling between the prior and data distribution. 



Most variants of VAE along these two directions require the closed-form KL divergence. This is essential in implementation but somehow limits the potential applications of more flexible pairs of posterior and prior, which may have no closed-form of KL divergence but could better capture the latent data distribution and alleviate model collapse. To improve the flexibility of the choice of posterior and prior in VAE, in this paper, we estimate the posterior by the expectation of kernel density estimator and derive a corresponding upper bound of KL-divergence, which has closed-form for many distributions. Such elasticity   makes the derivation of the optimal functional form of kernel possible and the implementation time efficient.  

It is true  we can simply employ Monte-Carlo simulation (MC) to approximate KL-divergence for complicated posteriors. But the price is the time efficiency. For example, \cite{Davidson2018HypersphericalVA} employed  von Mises-Fisher (vMF) distribution instead of Gaussian to better capture latent hyperspherical structure. With uniform prior, their proposed Hyperspherical Variational Auto-Encoders(HVAE) has closed form of KL-divergence while sampling from von Mises-Fisher (vMF) distribution requires acceptance-rejection method, which can be time consuming for high-dimensional latent space. Additionally, \cite{AVAE} directly employed a Gaussian kernel density estimator to estimate the posterior, which requires a large number of samples and is time consuming to achieve decent performance. Our main contributions can be summarized as follows.

\begin{enumerate} 
    \item  We  formulate the latent space learning process in VAE as a problem of kernel density estimation. Inspired by the results in \cite{Bickel} , we then model the posterior by the expectation of corresponding kernel density estimator and derive an upper bound of KL-divergence, which has closed-form for many distributions. Some asymptotic results are established as well. See details in section \ref{KDE} and \ref{Optimal kernel of VAE}
    \item After that, we utilize the conclusion from \cite{Bickel} and show that the derived upper bound of KL-divergence is tightest when we employ Epanechnikov kernel. The derivation connects a quadratic functional  with KL-divergence, which to our best knowledge, is the first attempt to bridge the concept of asymptotic distribution of KDEs with VAE. 
    \item  Thanks to the reparametrization trick, the implementation of Epanechnikov kernel in VAE is  straightforward. We conduct detailed comparisons between Epanechnikov VAE (EVAE) and Gaussian VAE in many benchmark image datasets under standard encoder-decoder structure. The extensive experiments not only demonstrate the superiority of EVAE over VAE and other baselines in the quality of reconstructed images, as measured by the FID score and Sharpness\cite{WAE} but also illustrate that EVAE has reasonable time efficiency compared to VAE.
\end{enumerate}

 The remaining sections of this paper are organized as following schema. In section \ref{preliminary}, we provide some preliminaries  involving VAE and kernel density estimations. In section \ref{Optimal kernel of VAE}, we show that Epanechnikov kernel gives the tightest upper bound in bounding the KL-divergence in the functional sense under appropriate conditions. Based on results in section \ref{Optimal kernel of VAE}, we propose the Epanechnikov VAE (EVAE) in section \ref{EVAE}.  The comparisons between EVAE, vanilla VAE and other baselines in  benchmark datasets are illustrated  in section \ref{Experiments}.  Section \ref{discussion} discusses few characteristics and limitations of EVAE and suggests some future directions. The broader impact statement is enclosed in Appendix \ref{Broader Impact Statement}. The pytorch codes for the implementation of  EVAE and experiments  are available in supplementary materials. All experiments are performed  on a laptop with 12th Gen Intel(R) Core(TM) i7-12700H (2.30 GHz) ,16.0 GB RAM and NVIDIA 4070 GPU.
 
\section{Preliminary}
\label{preliminary}

\subsection{VAE formulation}
Consider a dataset $\mathbf{X}=\{\mathbf{x}^{(i)}\}_{i=1}^{n}$ consists of $n$ i.i.d samples from space $\mathcal{X}$ whose dimension is $d$. In VAE, we assume that every observed data $\mathbf{x}^{(i)}=(x_{1}^{(i)},x_{2}^{(i)},...,x_{d}^{(i)})\in \mathcal{X}$ is  generated by a latent variable $\mathbf{z}^{(i)}\in \mathcal{Z}$ whose dimension is $p$. Then the data generation process can be summarized in 2 steps. It first produces a  variable $\mathbf{z}^{(i)}$ from some prior distributions $p(\mathbf{z})$.  Then given the value of $\mathbf{z}^{(i)}$, an observed value $\mathbf{x}^{(i)}$ is generated from certain conditional distribution $p_{\theta}(\mathbf{x}|\mathbf{z})$. We typically assume $d>p$  and likelihood $p_{\theta}(\mathbf{x}|\mathbf{z})$ are differentiable  distributions w.r.t $\mathbf{\theta}$ and $\mathbf{z}$. 
To maximize the empirical log likelihood log$p_{\theta}(\mathbf{x})$, we need to evaluate the integral in the form
\[
\text{log }p_{\theta}(\mathbf{x})=\text{log}\int p_{\theta}(\mathbf{x},\mathbf{z})d\mathbf{z}= \text{log} \int p_{\theta}(\mathbf{x}|\mathbf{z})p(\mathbf{z})d\mathbf{z}\,,
\]
which is intractable in most cases. Fortunately, we can instead maximizing its log evidence lower bound (ELBO) $\mathcal{L}$ with the help of Jensen's inequality:
\begin{equation}
\label{ELBO}
    \text{log }p_{\theta}(\mathbf{x})\geq    \underset{{\substack{ \mathbf{z}\sim q_{\phi(\cdot|\mathbf{x})}} }}{\mathbb{E}}[\text{log }p_{\theta}(\mathbf{x|\mathbf{z}})]-KL(q_{\phi}(\mathbf{z}|\mathbf{x})||p(\mathbf{z})) \,,
\end{equation}
where the RHS of inequality (\ref{ELBO}) is called evidence lower bound (ELBO) and $KL(\cdot||\cdot)$ represents the KL-divergence between two distributions.

The conditional likelihood $p_{\theta}(\mathbf{x|\mathbf{z}})$, approximate posterior $q_{\phi}(\mathbf{z}|\mathbf{x})$ and the prior distribution $p(\mathbf{z})$ can be chosen independently. For convenience, most applications of VAE employ Gaussian parametrization for all three likelihoods. Since we would like to investigate new forms of posterior and prior rather than the form of conditional likelihood $p_{\theta}(\mathbf{x|\mathbf{z}})$,  we   can assume  a multivariate Bernoulli or  Gaussian model w.r.t  $p_{\theta}(\mathbf{x|\mathbf{z}})$ for simplicity. For example, under multivariate Gaussian, we have $\text{log}p_{\theta}(\mathbf{x|\mathbf{z}})=\text{log}\mathcal{N}(\mathbf{x};g_{\theta}(\mathbf{z}),I)$. As to multivariate Bernoulli model, we have  $\text{log }p_{\theta}(\mathbf{x|\mathbf{z}})=\sum_{i=1}^{d}[x_{i}\text{log }(g_{\theta}(\mathbf{z})_{i})+(1-x_{i})\text{log }(1-g_{\theta}(\mathbf{z})_{i})]$ where $g_{\theta}(\mathbf{z}):\mathcal{Z}\to \mathcal{X}$ are typically neural-network parametrizations. 

To maximize ELBO, we now need to minimize the following target function for given data $\mathbf{x}$:
\begin{equation}
\label{Objective function}
     \underset{{\substack{ \mathbf{z}\sim q_{\phi(\cdot|\mathbf{x})}} }}{\mathbb{E}} \left[ -\text{log }p_{\theta}(\mathbf{x|\mathbf{z}})\right]+KL(q_{\phi}(\mathbf{z}|\mathbf{x})||p(\mathbf{z}))\,.
\end{equation}
The two terms in  equation (\ref{Objective function}) are named as ``reconstruction error'' and ``divergence'' or ``regularization term'', respectively. The ``divergence'' term  regularizes the mismatch between approximate posterior and prior distribution.

\subsection{Model the posterior as the expectation of  kernel density estimator}
\label{KDE}
A common assumption of latent space in VAE is the factorization of approximate posterior, i.e.  dimensions of latent space are independent with each other. Under this assumption, we only need to  consider the KDE formulation for posterior $q_{\phi}(\mathbf{z}|\mathbf{x})$ and prior $p(\mathbf{z})$ in one-dimensional case. Similar arguments can be applied to multi-dimensional cases by additivity of KL-divergence under independence. For the consistency of notations, we still use bold letter $\mathbf{x}$ and $\mathbf{z}$ to denote $x$ and $z$ in one-dimensional KDE. Given $Y_{1},...,Y_{m}$ be i.i.d random variables with a continuous density function $f$, \cite{Parzen,Rosenblatt} proposed kernel density estimate $f_{m}(y)$ for estimating $f(y)$ at a fixed point $y\in \mathbb{R}$:
\begin{equation}
\label{kernel estimate}
    f_{m}(y)=\frac{1}{mb(m)}\sum_{i=1}^{m}K\left[ \frac{y-Y_{i}}{b(m)}\right]=\frac{1}{b(m)}\int K\left[\frac{y-t}{b(m)} \right]dF_{m}(t) \, ,
\end{equation}
where $F_{m}$ is the sample distribution function, $K$ is an appropriate kernel function such that $\int K(y)dy=1$ and the positive number $b_{m}$, which typically relies on the sample size $m$, is called bandwidth such that $b(m)\to 0, mb(m)\to \infty$ as $m\to \infty$\footnote{We omit the limits of integrals if they are $-\infty$ to $\infty$.}.





On the other hand, with inequality $\text{log }(t)\leq t-1$ (for $t>0)$ we can bound  KL-divergence as follows: 

\begin{equation}
        \label{newKLbound}
\begin{split}
      KL(q||p)=\int q(\mathbf{z})\text{log }\frac{q(\mathbf{z})}{p(\mathbf{z})}d\mathbf{z}&\leq \int q(\mathbf{z})(\frac{q(\mathbf{z})}{p(\mathbf{z})}-1)d\mathbf{z}=\int \frac{(q(\mathbf{z})-p(\mathbf{z}))^2}{p(\mathbf{z})} d\mathbf{z}\,.
\end{split}
\end{equation}

Let the posterior $q_{\phi}(\mathbf{z}|\mathbf{x})=\mathbb{E}_{0}(q_{m,\phi}(\mathbf{z}))$, where 
\[
q_{m,\phi}(\mathbf{z})=\frac{1}{mb(m)}\sum_{j=1}^{m}K_{\phi,\mathbf{x}}\left[ \frac{\mathbf{z}-\mathbf{Z}_{j}}{b(m)}\right],\quad \mathbf{Z}_{j} \overset{i.i.d}{\sim} p(\Tilde{\mathbf{z}})\,.
\]

is a kernel density estimator of $q_{\phi}(\mathbf{z}|\mathbf{x})$ with kernel $K_{\phi,\mathbf{x}}$, given KDE sample size $m$, data point $\mathbf{x}$ and parameter $\phi$. Expectation $\mathbb{E}_{0}$ is taken w.r.t the samples from  prior distribution $p(\Tilde{\mathbf{z}})$. In other words, we model the posterior as the expectation of the kernel density estimator. The subscript $\phi$ of kernel function implies that the parameters in kernel can be learned by neural networks. 

By inequality (\ref{newKLbound}) and Jensen inequality, we obtain

\begin{equation}
\label{approximation_kernel}
\begin{split}
        KL(q_{\phi}(\mathbf{z}|\mathbf{x})||p(\mathbf{z}))&\leq \int \frac{[\mathbb{E}_{0}(q_{m,\phi}(\mathbf{z}))-p(\mathbf{z})]^2}{p(\mathbf{z})}d\mathbf{z} \leq \mathbb{E}_{0}\left[\int \frac{(q_{m,\phi}(\mathbf{z})-p(\mathbf{z}))^{2}}{p(\mathbf{z})}d\mathbf{z}\right]\,.
\end{split}
\end{equation}

To this end, we obtain a new lower bound of log-likelihood of data:

\begin{equation}
\label{new ELBO}
\begin{split}
        \text{log }p_{\theta}(\mathbf{x})&\geq  \mathbb{E}_{q_{\phi}(\mathbf{z}|\mathbf{x})}[\text{log }p_{\theta}(\mathbf{x}|\mathbf{z})]-KL(q_{\phi}(\mathbf{z}|\mathbf{x})||p(\mathbf{z}))
        \\
        &\geq \mathbb{E}_{q_{\phi}(\mathbf{z}|\mathbf{x})}[\text{log }p_{\theta}(\mathbf{x}|\mathbf{z})]-\mathbb{E}_{0}\left[\int \frac{(q_{m,\phi}(\mathbf{z})-p(\mathbf{z}))^{2}}{p(\mathbf{z})}d\mathbf{z}\right]\,.
\end{split}
\end{equation}

Since we model the posterior as the expectation of the kernel density estimator, the number of latent variable $Z$ generated from prior distribution, which is $m$,  is just used for theoretical derivation and we found that sampling  latent variable $Z$ once is good enough in practice, as suggested by \cite{VAE}.

In the kernel density estimation theory, \cite{Epanechnikov} showed that the Epanechnikov kernel minimizes $\int \mathbb{E}_{0}[q_{m,\phi}(\mathbf{z})-p(\mathbf{z})]^{2}d\mathbf{z}$ asymptotically, which gives us a hint of the functional optimization of kernel $K$ in right hand side of inequality (\ref{approximation_kernel}). The main assumptions of deriving optimal functional form of kernel are enclosed in Appendix \ref{assumptions}.
\section{Choice of kernel}

\label{Optimal kernel of VAE}
Let $f_{m}(t)$ be a kernel density estimate of a continuous density function $f$ at $t$, as defined in equation (\ref{kernel estimate}), we  construct a statistic $T_{m}$ as follows:
\[
T_{m}=mb(m)\int [f_{m}(t)-f(t)]^{2}a(t)dt\,,
\]
where $a(t)$ is an appropriate weight function. We now restate the main result of \cite{Bickel} as Theorem \ref{Theorem Bickel}:

\begin{theorem}[Bickel \& Rosenblatt \cite{Bickel}]
\label{Theorem Bickel}
    Let assumptions $\textbf{A1}-\textbf{A4}$ in Appendix \ref{assumptions} hold and suppose that the weight function a is integrable piecewise continuous and bounded. Suppose  $b(n)=o(n^{-\frac{2}{9}})$ and $o(b(n))=n^{-\frac{1}{4}}(\text{log}(n))^{\frac{1}{2}}(\text{log}\text{log}n)^{\frac{1}{4}}$ as $n\to \infty$, then $b^{-\frac{1}{2}}(n)(T_{n}-I(K)\int f(t)a(t)dt)$ is asymptotically normally distributed with mean 0 and variance $2J(K)\int a^{2}(t)f^{2}dt$ as $n\to \infty$, where 
    \begin{equation}
    \label{I(K)}
       I(K)=\int K^{2}(t)dt,\quad J(K)=\int \left[ \int K(t+y)K(t)dt\right]^{2}dy\,.     
    \end{equation}

\end{theorem}

In other words, under Theorem \ref{Theorem Bickel}, we have 
\[
\mathbb{E}[T_{m}]\to I(K)\int f(t)a(t)dt \quad \text{, as $m\to \infty$}\,.
\]

Let $mb(m)/n=1$ and $b(m)$ satisfies the conditions in Theorem \ref{Theorem Bickel},  by the asymptotic result of Theorem \ref{Theorem Bickel}, the inequality (\ref{approximation_kernel}) becomes\footnote{For the consistency of notations, we still use bold $\mathbf{x}$ and $\mathbf{z}$ to denote one-dimensional variable $x$ and $z$ in inequality (\eqref{Upperbound KL}) under the context of Theorem \ref{Theorem Bickel}.}:
\begin{equation}
    \begin{split}
    \label{Upperbound KL}
           KL(q_{\phi}(\mathbf{z}|\mathbf{x})||p(\mathbf{z}))&\leq  \frac{mb(m)}{n}\mathbb{E}_{0}\left[\int \frac{(q_{m,\phi}(\mathbf{z})-p(\mathbf{z}))^{2}}{p(z)}d\mathbf{z}\right]\\
           &=\frac{1}{n}\mathbb{E}_{0}\left[mb(m)\int \frac{(q_{m,\phi}(\mathbf{z})-p(\mathbf{z}))^{2}}{p(\mathbf{z})}d\mathbf{z}\right]\\
           &\overset{\text{n large}}{\approx}\frac{I(K_{\phi,\mathbf{x}})\int p(\mathbf{z})\frac{1}{p(\mathbf{z})}d\mathbf{z} }{n}=B\frac{I(K_{\phi,\mathbf{x}})}{n} \,,
    \end{split}
\end{equation}

where B is the length of  support interval of prior $p(\mathbf{z})$. For simplicity, we assume all data points have the same length of  support interval. Note that if the prior has infinite length of support, the inequality (\ref{Upperbound KL}) becomes theoretically useless. However, in practice, we found that parameter B is still useful for Gaussian prior. See Appendix \ref{EVAE with gaussian prior}. We can now find the kernel which gives the tightest upper bound of KL-divergence i.e. the kernel $K^{*}_{\phi,\mathbf{x}}$ minimizing $I(K_{\phi,\mathbf{x}})$  given  fixed parameters and data point $\mathbf{x}$.  Lemma \ref{Optimal} shows that Epanechnikov kernel  is the optimal choice.

\begin{lemma}
    \label{Optimal}
    Let $\mathcal{K}$ be the set of all $L^{1}(-\infty,\infty)$ nonnegative functions $K$ satisfying
    \[
   \int K(t)dt=1, \int tK(t)dt=\mu, \int (t-\mu)^2K(t)dt=\frac{1}{5}r^{2}
    \] where $\mu \in (-\infty,\infty),r>0$. Then the functional $I(K)$ in equation (\ref{I(K)}) is minimized on $\mathcal{K}$ uniquely by 
    \[
    K^{*}(t)=\begin{cases} 
     \frac{3}{4r}\left(1-\left(\frac{t-\mu}{r}\right)^2\right) & t \in[\mu-r,\mu+r] \\
      0 & \text{otherwise} \\
      \end{cases}
    \] and $min_{K}I(K)=I(K^{*})=\frac{3}{5r}$. The optimal kernel $K^{*}$ is named as Epanechnikov kernel \citep{Epanechnikov}.
\end{lemma}

The proof is based on the idea of Lagrange multiplier. See Appendix \ref{proof of lemma 2}.



\section{Epanechnikov VAE}
\label{EVAE}
Getting inspired by the functional optimality of Epanechnikov kernel in controlling KL-divergence, we propose the Epanechnikov Variational Autoencoder (EVAE) whose resampling step is based on Epanechnikov kernel. There are two main differences between EVAE and VAE. The latent distribution  in EVAE is assumed to be estimated by the Epanechnikov kernel rather than multivariate isotropic Gaussian. And EVAE is trained to minimize a different target function (\ref{Target function EVAE}):
\begin{equation}
    \label{Target function EVAE}
       \underset{{\substack{z\sim q_{\phi(\cdot|\mathbf{x})}} }}{\mathbb{E}} \left[ -\text{log }p_{\theta}(\mathbf{x|\mathbf{z}})\right]+B\frac{I(K^{*}_{\phi,\mathbf{x}})}{n}\,,
\end{equation}
where $n$ can be the sample size or minibatch size, $\phi$ are outputs of the encoding network, $K^{*}_{\phi,\mathbf{x}}$ is Epanechnikov kernel given data point $\mathbf{x}$ and trainable neural network parameter $\phi$, and support parameter $B$ is a constant. Note that target function (\ref{Target function EVAE}) is an upper bound of equation (\ref{Objective function}) used in ordinary VAE. Suppose we have $M$ data points in each minibatch, the sample version of equation (\ref{Target function EVAE}) at $i$-th data point $\mathbf{x}^{(i)}$ is
\begin{equation}
\label{sample ELBO}
  \tilde{\mathcal{L}}(\mathbf{\theta},\mathbf{\phi},\mathbf{B};\mathbf{x}^{(i)})\approx -\frac{1}{L}\sum_{l=1}^{L}(\text{log }p_{\theta}(\mathbf{x}^{(i)})|\mathbf{z}^{(i,l)})+\frac{3B}{5M}\sum_{k=1}^{p}\frac{1}{r_{k}^{(i)}}  \,,
\end{equation}
where $i\in\{1,2,...,M\}$, $p$ is the dimension for latent space, B is the support length of prior (which we set to be constant for each hidden dimension and all data points), $\mathbf{\mu}^{(i)}=(\mu_{1}^{(i)},...,\mu_{p}^{(i)}), \mathbf{r}^{(i)}=(r_{1}^{(i)},...,r_{p}^{(i)})$\footnote{The encode network output $\mathbf{\mu}^{(i)}$ is used to sample from Epanechnikov based posterior and only optimized in the reconstruction term.} are outputs of the encoding networks with variational parameters $\mathbf{\phi}$ and $M$ is minibatch size. In vanilla VAE, \cite{VAE} suggested that the number of regenerated samples $L$ can be set to 1 as long as the minibatch size is relatively large. Based on our experiment experience, this is also the case for EVAE. To sample latent variables from the  posterior, we need to derive the density of $ q_{\phi}(\mathbf{z}|\mathbf{x}^{(i)}) $:
\begin{equation}
\label{approx posterior}
\begin{split}
        q_{\phi}(\mathbf{z}|\mathbf{x}^{(i)}) &= \mathbb{E}_{0}\left[\frac{1}{mb(m)}\sum_{j=1}^{m}K^{*}_{\phi,\mathbf{x}^{(i)}}\left( \frac{\mathbf{z}-\mathbf{Z}_{j}}{b(m)}\right) \right]\\
        &=\mathbb{E}_{0}\left[\frac{1}{b(m)}K^{*}_{\phi,\mathbf{x}^{(i)}}\left( \frac{\mathbf{z}-\mathbf{Z}_{1}}{b(m)}\right) \right] \\
        &=\int \frac{1}{b(m)}K^{*}_{\phi,\mathbf{x}^{(i)}}\left( \frac{\mathbf{z}-\mathbf{\Tilde{z}}}{b(m)}\right) p_{\mathbf{z}}(\mathbf{\Tilde{z}})d\mathbf{\Tilde{z}}\,,
\end{split}
\end{equation}
where $p_{\mathbf{z}}$ is the density of prior distribution. Given a data point $\mathbf{x}^{(i)}$, let $ q_{\phi}(\mathbf{z}|\mathbf{x}^{(i)}) $ be the density of  random variable $Z(\mathbf{x}^{(i)})$ ,  $K^{*}_{\phi,\mathbf{x}^{(i)}}$ be the density of random variable $K^{*}(\mathbf{x}^{(i)})$ and  prior $p_{\mathbf{z}}$ is density of the random variable $Z$, then it's clear to see that the posterior is now a convolution between two random variables, i.e.
\begin{equation}
\label{decomposition}
    Z(\mathbf{x}^{(i)})=b(m)K^{*}(\mathbf{x}^{(i)})+Z\,.
\end{equation}

Essentially, identity (\ref{decomposition}) decomposes the posterior into two parts. One is the prior information,  the other  represents the incremental updates from new information induced by data $\mathbf{x}^{(i)}$. The Epanechnikov kernel can be viewed as the ``optimal'' direction of perturbing the prior distribution and the coefficient $b(m)$ can be interpreted as ``step size'', which controls the deviation of posterior from the prior. The resampling step in EVAE can be divided into two parts as well, as described in Algorithm \ref{EVAE sample} where we used uniform distribution as the prior in EVAE. For the sake of finite support and simplicity, we assume that the prior distribution is uniformly distributed and step size $b(m)$ is the same for all dimensions. The theoretical conditions for $b(m)$ is demanding. In practice we found that  setting coefficient $b(m)$ as $b(100)=100^{-2/9}\approx 0.3594 $ is good enough.
 
   \begin{algorithm}

        \caption{Resampling step in a minibatch of EVAE}
  \label{EVAE sample}        
        \begin{algorithmic}[1]
            \Require Latent space dimension $d_{z}$; Minibatch size $M$. Step size $b(m)$. Prior support interval length $B$.  Mean $\mathbf{\mu}_{\phi}(\mathbf{x})$ , spread  $\mathbf{r}_{\phi}(\mathbf{x})$  are all learned by encoder networks given input $\mathbf{x}$ and have dimension $M\times d_{z}$.
            \State Sample a  $M\times d_{z}$ matrix $\mathbf{U}$  where each $(i,j)$ entry of the random matrix $\mathbf{U}$ is sampled from $\text{Unif}[-B/2,B/2]$.          
            \State Sample a $M\times d_{z}$ matrix $\mathbf{K}$ where each $(i,j)$ entry of the random matrix $\mathbf{K}$ is sampled from an standard Epanechnikov kernel supported on $[-1,1]$.
            \State Shift and scale sampled $\mathbf{K}$ by the location-scale formula: $\mathbf{Z}=\mathbf{\mu}_{\phi}(\mathbf{x})+\mathbf{r}_{\phi}(\mathbf{x}) \odot \mathbf{K}$.
            
            \State  \textbf{Return:} $b(m)\odot \mathbf{Z}+\mathbf{U}$
        \end{algorithmic}
    \end{algorithm}


    \begin{algorithm}

        \caption{Sampling from centered Epanechnikov kernel supported on $[-1,1 ]$}
          \label{Sample from EK}
        \begin{algorithmic}[1]
            \State Sample $U_{1},U_{2},U_{3}\overset{i.i.d}{\sim} \text{Unif}[-1,1]$.
            \State Set $U=\text{Median}(U_1,U_2,U_3)$
            \State  \textbf{Return:} $U$
        \end{algorithmic}
    \end{algorithm}


In Algorithm \ref{EVAE sample}, we apply reparametrization trick to sample $\mathbf{z}$ from general Epanechnikov kernel, i.e $\mathbf{z}^{(i,l)}=\mathbf{\mu}^{(i)}+\mathbf{r}^{(i)} \odot \mathbf{k}^{(l)}$, where $\mathbf{k}^{(l)}$ is sampled from   standard  Epanechnikov kernel supported on $[-1, 1]$  as it lies in the "location-scale" family. We use $\odot$ to signify the element-wise product.  There are many ways to sample from standard Epanechnikov kernel, such as accept-rejection method. For efficiency, we provide a faster sampling procedure in Algorithm \ref{Sample from EK}. The theoretical proof  is given in Appendix \ref{Theoretical support for Algorithm}. 


\section{Experiments}

\label{Experiments}

In this section, we will compare the proposed EVAE model with (vanilla) VAE whose posterior and prior are modelled by isotropic multivariate Gaussian in real datasets. To assess the quality of reconstructed images, we employ the Frechet Inception Distance (FID) score \cite{FID} to measure the distribution of reconstructed images with the distribution of real images. The lower, the better.  Inception\_v3 \cite{Inceptionv3} is employed as default model to generate features of input images, which is a standard implementation in generative models. Unlike the unbounded support of Gaussian distribution, the compact support for Epanechnikov kernel could help EVAE generate less blurry  images. To check this claim empirically, we  calculated sharpness \cite{WAE} for reconstructed images by a $3\times 3$ Laplace filter. See more details in Appendix \ref{Datasets and Training details}. 
\subsection{Benchmark datasets}
\begin{table}[H]
  \caption{VAE and EVAE results on CIFAR10 and CelebA datasets}
  \label{table cifar}
  \centering
 \begin{tabular}{lllll llll llll}
\toprule
 Datasets&\multicolumn{4}{c}{CIFAR-10}
&
\multicolumn{4}{c}{CelebA} \\
\cmidrule(l){2-5}\cmidrule(l){6-9}
\multirow{2}{*}{$d_{z}$} &\multicolumn{2}{c}{VAE} &\multicolumn{2}{c}{EVAE}    & \multicolumn{2}{c}{VAE} &\multicolumn{2}{c}{EVAE}      \\
 &  FID  &Sharpness &FID  &Sharpness    & FID  &Sharpness &FID&Sharpness   \\
\midrule

8 & 230.1& 0.0353&226.6&0.0337&198.7&0.0147&198.7&0.0154  \\
16 &  181.4& 0.0350& 164.2&0.0355 &152.5&0.0148& 149.7&0.0154      \\
32  & 149.7 &0.0348& 123.5&0.0359&   110.1& 0.0156&97.5& 0.0162\\
64 &  144.7&0.0355& 79.7& 0.0369&  77.2&0.0163&65.3& 0.0167\\
\bottomrule
\end{tabular}
\end{table}
We trained EVAE and VAE on four benchmark datasets: MNIST \cite{deng2012mnist}, Fashion-MNIST \cite{fasionmnist}, CIFAR-10 \cite{cifar10} and CelebA \cite{celeba}. The detailed information for training parameters  are attached in Appendix \ref{Datasets and Training details}. In terms of FID score, we observe that EVAE has an edge in high dimensions ($d_{z}=32,64$), indicating the better quality of reconstructed images from EVAE. Additionally, when $d_{z}=64$, EVAE generates higher sharpness of reconstructed images for all datasets, as illustrated in Table \ref{table cifar}. See more results in Appendix \ref{Benchmark datasets}. This result empirically justifies the positive effect of having compact support in posteriors and priors.  As to  CIFAR-10, EVAE has larger sharpness in high dimensions ($d_{z}=32,64$) while it is relatively mediocre in low dimensions. Possible reasons include low-resolution(blurriness) of original images and deficient expressibility  of simple CNN models in low dimensions.  But the  validation reconstruction loss curves in Appendix \ref{loss curve}  indeed authenticate the superiority of EVAE over benchmark datasets, including CelebA.


\subsection{Reconstruction samples}

\captionsetup[figure]{font=small}
\begin{figure}[H]
\subfloat[Real images]{\label{control_prob} \includegraphics[width=0.32\columnwidth]{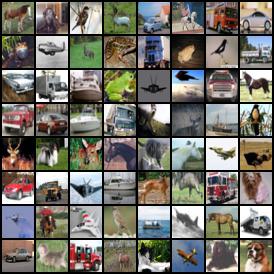}}
    \hfill
    \subfloat[VAE reconstructed images]{\label{small_signal} \includegraphics[width=0.32\columnwidth]{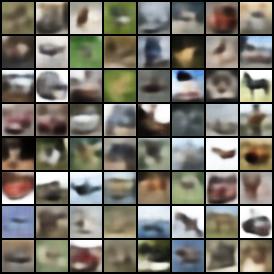}}
    \hfill
\subfloat[EVAE reconstructed images]{\label{mp} \includegraphics[width=0.32\columnwidth]{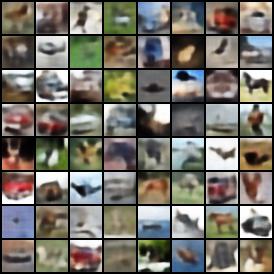}}
\caption{(a) Sampled real images from hold-out samples in CIFAR-10 (b) Reconstructed images by VAE. (c) Reconstructed images by EVAE. Dimension $d_{z}=64$ for both models.  }
\label{cifar10 sample}
\end{figure}
 Figure \ref{cifar10 sample} presents  some test reconstructed samples  from trained VAE and EVAE with $d_{z}=64$ of CIFAR10, respectively. We can see many images reconstructed from EVAE are able to pick up local features better than the VAE. And reconstructed samples from  CIFAR-10 are clearer  and closer to  the original images, as indicated by large  gap between FID scores. See more examples of MINST, Fashion-MNIST and CelebA-64 datasets in Appendix \ref{Sampled reconstructed images}.   The statistical analysis from binomial test based on total  experiments (Appendix \ref{Datasets and Training details}) shows that EVAE significantly outperforms VAE in FID and sharpness.



\subsection{Unconditional samples}

\begin{figure}[H]
\subfloat[VAE]{\label{vaeprior} \includegraphics[width=0.22\columnwidth]{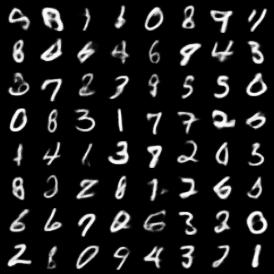}}
    \hfill
    \subfloat[EVAE(B=0.1)]{\label{evae_prior0.1} \includegraphics[width=0.22\columnwidth]{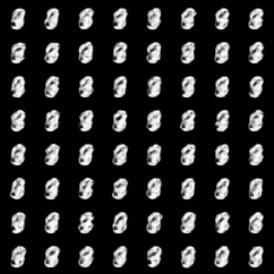}}
    \hfill
\subfloat[EVAE(B=1)]{\label{evae_prior1} \includegraphics[width=0.22\columnwidth]{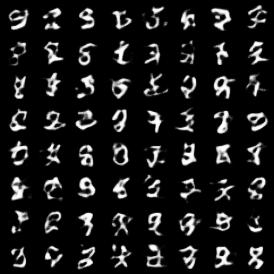}}
\hfill
\subfloat[EVAE(B=10)]{\label{evae_prior10} \includegraphics[width=0.22\columnwidth]{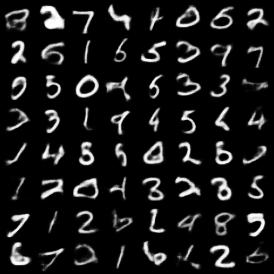}}
\caption{(a) Unconditional samples  generated from VAE (MNIST dataset) (b) Unconditional samples generated from EVAE with B=0.1 (c) Unconditional samples generated from EVAE with B=1 (d) Unconditional samples generated from EVAE with B=10  }
\label{prior sample evae}
\end{figure}

Figure \ref{prior sample evae} demonstrates the unconditional samples from MINST dataset with EVAE \footnote{We employed uniform prior in the rest of experiments. Appendix \ref{EVAE with gaussian prior} already showed that there is no big difference for the model performance between uniform prior and Gaussian prior. The training details follows Appendix \ref{Model architecture} and \ref{Datasets and Training details}.} and VAE  and different values of parameter B. In other words, we directly  sample from prior $p(z)$, multiply it by B and feed latent samples into decoder to obtain the unconditional samples. We  observe that  the larger the value of B is, the more diversed novel samples will be generated   from EVAE. This makes sense as smaller value of B will make the model focus more on the perturbation part (KDE based) of identity (\ref{decomposition}), leading to the prior collapse. To this point, we can view the parameter B as a way to trade off the reconstruction and sample diversity.

\captionsetup[figure]{font=small}

\subsection{Comparisons with baselines}

In this section we conducted extra experiments comparing EVAE with open-source variants of VAE that improve the representational capacity of the encoder such as $\beta-$VAE \citep{Higgins2016betaVAELB}, IWAE \citep{Burda2015ImportanceWA}, WAE \citep{WAE} and HVAE \citep{Davidson2018HypersphericalVA} on three datasets. Note that we didn't involve normalizing flow in the baselines as it belongs to a different type of image generation models.  We employed uniform prior in EVAE and we ran all experiments for three times. We also reported mean and standard deviation of FID score (for testing set reconstruction quality). Same for sharpness. For fair comparisons, we used same encoder and decoder CNN architecture for all baselines. The only difference come from posterior modeling and prior modeling. All latent dimension $d_{z}$ are set to be 64. Other hyperparameters and training details are the same with details in Appendix \ref{Model architecture} and \ref{Datasets and Training details}. 

From Table \ref{baseline}, we observe that EVAE with $B=0.1$ performs universally good in FID and sharpeness. Our support length parameter $B$ plays the similar role with $\beta$ in $\beta-$VAE, which verifies the intuition in the discussion section. In other words, $B$ and $\beta$ somehow control the reconstruction quality and sample diversity. For other baselines, WAE(MMD) performs also well in many cases, which is expected as it generalizes the KL divergence to the Wasserstein distance. IWAE  has decent performance by utilizing importance weight sampling to shrink the gap between ELBO and log-likelihood, where $k=5$ represents the number of samples in the importance sampling step. Hyperspherical VAE(HVAE) seems mediocre in our experiments, which might be due to the complexity of von-Mises Fisher prior and choice of dimension of hypershere manifold. All the  methods  shared similar running time except for HVAE  as it requires acceptance-rejection method for the sampling step, which is relatively time consuming for high-dimensional case.

\begin{table}[H]
  \caption{EVAE and other baselines }
  \label{baseline}
  \centering
 \begin{tabular}{l l lll}

\cmidrule(l){1-5}
\multirow{2}{*}{Models} &\multicolumn{2}{c}{Fashion-MNIST}    & \multicolumn{2}{c}{CIFAR10}       \\
&FID  &Sharpness    & FID  &Sharpness  \\
\midrule

VAE & 39.1(0.3)& 0.016(3E-4)&142.9(1.7)&0.035(7E-4)\\
EVAE($\text{B}=0.1$) &\textbf{12.2(0.2)}& \textbf{0.027(2.1E-4)}&\textbf{79.7(0.1)}&\textbf{0.036(7E-4)} \\
EVAE($\text{B}=10$) & 32.6(0.5)& 0.019(4E-4)&154.8(2.6)& 0.035(6E-4)\\
$\beta$-VAE($\beta=0.1$) & 19.5(0.5)& 0.023(5E-4)&  84.4(0.8)&0.035(5E-4)\\
$\beta$-VAE($\beta=5$) & 74.5(2.0) &0.01(2E-4)&  239.8(1.9)&0.034(1.4E-3)\\
IWAE($k=5$) & 26.8(0.1)& 0.02(3E-4)&  125.4(1.6)&0.036(7E-4)\\
WAE(MMD) & 22.1(0.5)& 0.021(6E-4)&  140.1(7.2)&0.034(7E-4)\\
HVAE & 36.8(0.3)& 0.016(1E-4)&  195.2(0.8)&0.035(1E-3)\\
\bottomrule
\end{tabular}
\end{table}

\subsection{Extra experiments}
 For completeness, we conducted extra experiments to see the performance of EVAE with Gaussian prior, effect of uniform prior support parameter B  and the time efficiency of EVAE in Appendix \ref{More results}. In short, Appendix \ref{EVAE with gaussian prior} implies that the choice of prior doesn't effect the performance of EVAE very much. Appendix \ref{effect of B} shows that the support pamameter B plays an role in controlling the  reconstruction quality. And Table \ref{time efficiency table} in Appendix \ref{time efficiency} verifies that EVAE is time efficient due to its simple implementation.




\section{Discussion and limitation}

\label{discussion}



\subsection{Connections between EVAE and $\beta$-VAE}

From target function (\ref{sample ELBO}), we can view the ratio $\frac{B}{M}$ can  as a weight parameter in penalizing the kernel term. This formula is similar to the ELBO given in $\beta$-VAE. It would be interesting to compare the  disentanglement and implicit regularization effect \citep{ImplicitRegularization} of EVAE and $\beta$-VAE. In fact, in EVAE the weight parameter $\frac{B}{M}$ is derived from the global deviation result for KDEs, i.e. Theorem \ref{Theorem Bickel}, where the constant B should be interpreted as the length of support interval for the prior. 




On the other hand, the limitations of EVAE involve following few points:

\subsection{A more precise approximation of KL-divergence}


In section \ref{KDE}, we bounded the KL-term by a simple inequality $\text{log }t \leq 1-t$, which limits the potential of EVAE in more complicated cases such as high resolution images. It's possible to derive sharper bounds with higher order approximation where the Epanechnikov kernel may not be the optimal one. 

\subsection{Optimal kernel under different criterions}

In this paper, we mainly focused on the $L_{2}$ deviation of KDEs, which is measured by the functional $I(K)$ proposed in Theorem \ref{Theorem Bickel}. However, in the general theory of KDEs, different criterions lead to different optimal kernels. For example, we didn't put too much attention on the convolution functional $J(K)=\int \left[ \int K(t+y)K(t)dt\right]^{2}dy $, which is related to the asymptotic variance of statistic $T_{m}$ defined in section \ref{Optimal kernel of VAE}. If we want to minimize the asymptotic variance of  $T_{m}$, the optimal kernel is just the uniform kernel, as derived by \cite{Huang}. 

In addition, the Bickel-Rosenblatt statistic $T_{m}$ we defined in section \ref{Optimal kernel of VAE} is proposed by \cite{Bickel}, which can be viewed as a ``continuous version'' of the ordinary k-cell Chi-Square statistic and the KL-divergence is also a type of Chi-Square function \cite{MinimumChiSquare}. Replacing KL-divergence with expectation of another type of Chi-Square statistic makes the kernel optimization (minimum Chi-Square estimation) in section \ref{Optimal kernel of VAE}  possible, which supports the claim in \cite{MinimumChiSquare} that minimum Chi-Square estimate is more general and flexible.



\bibliography{ref}
\bibliographystyle{plain}
\newpage
\appendix
\section*{Appendix}

\section{Assumptions for Theorem \ref{Theorem Bickel}}
\label{assumptions}
We mainly borrow the terminologies and assumptions in \cite{Rosenblatt} which is the  pioneering work in measuring deviations of density function estimates. Note that \cite{Rosenblatt} studied the asymptotic distribution of the quadratic functional $\int [f_{n}(t)-f(t)]^{2}a(t)dt$ under appropriate weight function $a$  and conditions as sampling size $n$ approaches to infinity. In inequality (\ref{approximation_kernel}), we just saw that the weight function is the reciprical of prior density in our case. Assumptions $\textbf{A1}-\textbf{A4}$ are listed as follows:

\begin{enumerate}[start=1,label={(\bfseries A\arabic*):}] 
\item The kernel function $K$ is bounded, integrable, symmetric (about 0) and $\int K(t)dt=1, \int t^2K(t)dt<\infty, \int K^{2}(t)dt<\infty$. Also, $K$ either (a) is supported on an closed and bounded interval $[-B,B]$ and is absolutely continuous on $[-B,B]$ with derivative $K'$ or (b) is absolutely continuous on the whole real line with derivative $K'$ satisfying $\int |K'(t)|^{k}dt<\infty,k=1,2$.  Moreover, 
\[
\int_{t\geq 3} |t|^{\frac{3}{2}}[\text{log }(\text{log }|t|)]^{\frac{1}{2}}[|K'(t)|+|K(t)|]dt<\infty
\]
\item The underlying density $f$ is continuous, positive and bounded.
\item Squared density $f^{1/2}$ is absolutely continuous and its derivative is bounded in absolute value. 
\item The second derivative $f''$ exists and is bounded.
\end{enumerate}

We can see that the Gaussian kernel  satisfies those assumptions, indicating that the kernel density estimation theory also works for Gaussian VAE. Similarly, most priors and posteriors we listed in section \ref{introduction} lie in conditions  $\textbf{A1}-\textbf{A4}$, demonstrating the promising applications of kernel estimate theory in variants of VAE.

\section{Proofs}

\subsection{Proof of Lemma \ref{Optimal}}
\label{proof of lemma 2}
For simplicity, we first consider the following  constraints 

\[
K\geq 0, \quad \int K(t)dt=1,\quad \int tK(t)dt=0, \int t^2K(t)dt=\frac{1}{5}\,.
\]

By the method of undermined multipliers, it's equivalent to minimize the following target functional without constraints:

\[
\int K^{2}(t)+aK(t)+ct^2K(t)dt\,,
\]
with simplified constraints above and $a,c$ are undermined real coefficients. We ignore the term $tK(t)$ as it does not contribute to the unconstrained target function now.

For fixed $t$, denote $y(K)=K^2+aK+ct^2K, (K\geq 0)$. Note that the quadratic function $y(K)$ achieves minimum when $K=-\frac{c}{2}t^2-\frac{a}{2}$.

It follows that $y(K)$ is minimized subject to $K\geq 0$ by
\[
K(t)=\begin{cases} 
-\frac{c}{2}t^2-\frac{a}{2} & -\frac{c}{2}t^2-\frac{a}{2}\geq0\,. \\
0 & -\frac{c}{2}t^2-\frac{a}{2}<0\,.
   \end{cases}
\]

We can rewrite it as 
\[
K(t)=\begin{cases} 
A(B^2-t^2) & |t|\leq B \,.\\
0 & \text{otherwise}\,.
   \end{cases}
\]
for some number $A,B$. By simplified assumptions $\int tK(t)dt=0, \int t^2K(t)dt=\frac{1}{5}$, we can find that $A=\frac{3}{4},B=1$. 

Under general moment conditions in Lemma \ref{Optimal}, optimal $K^{*}$ can be written as

\[
K^{*}(t)=\begin{cases} 
\frac{3}{4r}\left(1-\left(\frac{t-\mu}{r}\right)^2\right) & |t-\mu|\leq r \,.\\
0 & \text{otherwise}\,.
   \end{cases}
\]

by location-scale formula. In literature \cite{Epanechnikov}, this kernel is called Epanechnikov kernel. We put $1/5$ in front of the constraint of second moment in order to make the resulting support interval cleaner, which won't change the optimal kernel.  The corresponding optimal value of $I(K)$ is $I(K^{*})=\frac{3}{5r}$.

\textbf{Plots of Standard Epanechnikov kernel and Guassian kernel}
\renewcommand{\thefigure}{S\arabic{figure}}
\captionsetup[figure]{font=small}
\begin{figure}[H]
\centering
\includegraphics[width=0.7\columnwidth]{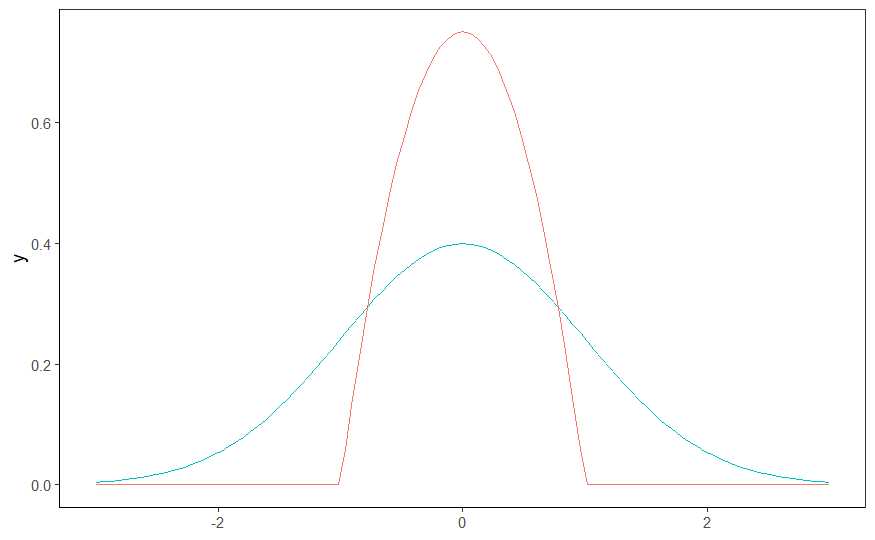}
\caption{Red curve: Standard Epanechnikov kernel. Green curve: Standard Gaussian kernel. }
\label{std E kernel vs std G kernel}
\end{figure}

\subsection{Theoretical support for Algorithm \ref{Sample from EK}}
\label{Theoretical support for Algorithm}
Given $U_1,U_2,U_3 \overset{i.i.d.}{\sim} \text{Unif}[-1,1]$, we only need to show the density of median$(U_1,U_2,U_3)$  is standard Epanechnikov kernel.

Denote $Y=\text{Median}(U_1,U_2,U_3)$, we have

\begin{equation*}
    \begin{split}
        P(Y\leq t)&=P(\text{Median}(U_1,U_2,U_3)\leq t)\\
        &=P(\text{Exactly two of $U_{1},U_2,U_3$ are less than $t$} )+P(\text{All of $U_{1},U_2,U_3$ are less than $t$})\\
        &=\binom{3}{2}\left( \frac{1+t}{2}\right)^{2}\left( \frac{1-t}{2}\right)+\binom{3}{3}\left( \frac{1+t}{2}\right)^{3}\\
        &=\frac{1}{2}+\frac{3}{4}t-\frac{t^3}{4}\,.
    \end{split}
\end{equation*}

Then the density $f_{Y}(t)$ of $Y$ is 
\[
f_{Y}(t)=\frac{3}{4}-\frac{3}{4}t^2=\frac{3}{4}(1-t^2)\,,
\]
which is essentially the standard Epanechnikov kernel.

\section{Model architecture}
\label{Model architecture}
\subsection{MNIST and Fashion-MNIST}

We used fully convolutional architectures with $4\times 4$ convolutional filters for both encoder and decoder in EVAE and VAE, as described following. All convolutions in the encoder and decoder employed SAME padding.

We  resized images in MNIST and Fashion-MNIST from $28\times 28$ to $32\times 32$ at beginning. In the last conv layer, the sigmoid activation function was used to restrict the range of output as we assumed Bernoulli type model of $p_{\theta}(\mathbf{x}|\mathbf{z})$ and the binary cross entropy loss employed used (reduction to sum). Dimensions $d_{z}$ for latent space : $\{8,16,32,64\}$

\textbf{Encoder $q_{\phi}$:}

\begin{equation*}
    \begin{split}
        x\in \mathbb{R}^{32\times 32} &\to \text{32 Conv, Stride 2} \to \text{BatchNorm} \to \text{ReLU}\\
        &\to \text{64 Conv, Stride 2} \to \text{BatchNorm} \to \text{ReLU}\\
        &\to \text{128 Conv, Stride 2} \to \text{BatchNorm} \to \text{ReLU}\\
        &\to \text{256 Conv, Stride 2} \to \text{BatchNorm} \to \text{ReLU}\\
        &\to \text{Fully connected ($1*1*256 \times d_{z}$) for each parameters}
    \end{split}
\end{equation*}

\textbf{Decoder $p_{\theta}$:}

\begin{equation*}
    \begin{split}
        z\in \mathbb{R}^{d_{z}\times d_{z}} &\to \text{Fully connected ($ d_{z}\times 1*1*256  $)}\\      
        &\to \text{128 ConvTran, Stride 1} \to \text{BatchNorm} \to \text{ReLU}\\
        &\to \text{64 ConvTran, Stride 2} \to \text{BatchNorm} \to \text{ReLU}\\
        &\to \text{32 ConvTran, Stride 2} \to \text{BatchNorm} \to \text{ReLU}\\
        &\to \text{1 ConvTran, Stride 2} \to \text{Sigmoid}
    \end{split}
\end{equation*}

\subsection{CIFAR-10}

Again, we used fully convolutional architectures with $4\times 4$ convolutional filters for both encoder and decoder in EVAE and VAE for CIFAR-10 model. In encoder, we employed a layer of Adaptive Average pool filter.  Other settings are the same with MNIST and Fashion-MNIST.

\textbf{Encoder $q_{\phi}$:}

\begin{equation*}
    \begin{split}
        x\in \mathbb{R}^{32\times 32} &\to \text{32 Conv, Stride 2} \to \text{BatchNorm} \to \text{ReLU}\\
        &\to \text{64 Conv, Stride 2} \to \text{BatchNorm} \to \text{ReLU}\\
        &\to \text{128 Conv, Stride 2} \to \text{BatchNorm} \to \text{ReLU}\\
        &\to \text{256 Conv, Stride 2} \to \text{BatchNorm} \to \text{ReLU}\\
        &\to \text{AdaptiveAvgPool2d}\\
        &\to \text{Fully connected ($1*1*256 \times d_{z}$) for each parameters}
    \end{split}
\end{equation*}

\textbf{Decoder $p_{\theta}$:}

\begin{equation*}
    \begin{split}
        z\in \mathbb{R}^{d_{z}\times d_{z}} &\to \text{Fully connected ($ d_{z}\times 1*1*256  $)}\\      
        &\to \text{128 ConvTran, Stride 1} \to \text{BatchNorm} \to \text{ReLU}\\
        &\to \text{64 ConvTran, Stride 2} \to \text{BatchNorm} \to \text{ReLU}\\
        &\to \text{32 ConvTran, Stride 2} \to \text{BatchNorm} \to \text{ReLU}\\
        &\to \text{3 ConvTran, Stride 2} \to \text{Sigmoid}
    \end{split}
\end{equation*}

\subsection{CelebA}

For CelebA dataset, we used  $5\times 5$ convolutional filters for both encoder and decoder in EVAE and VAE. Simiar to CIFAR-10,  we employed a layer of Adaptive Average pool filter before the fully connected layer in encoder. We first scaled images with Center Crop to $140\times 140$ and resized them to $64\times 64$.

\textbf{Encoder $q_{\phi}$:}

\begin{equation*}
    \begin{split}
        x\in \mathbb{R}^{64\times 64} &\to \text{64 Conv, Stride 2} \to \text{BatchNorm} \to \text{ReLU}\\
        &\to \text{128 Conv, Stride 2} \to \text{BatchNorm} \to \text{ReLU}\\
        &\to \text{256 Conv, Stride 2} \to \text{BatchNorm} \to \text{ReLU}\\
        &\to \text{512 Conv, Stride 2} \to \text{BatchNorm} \to \text{ReLU}\\
        &\to \text{AdaptiveAvgPool2d}\\
        &\to \text{Fully connected ($1*1*512 \times d_{z}$) for each parameters}
    \end{split}
\end{equation*}

\textbf{Decoder $p_{\theta}$:}

\begin{equation*}
    \begin{split}
        z\in \mathbb{R}^{d_{z}\times d_{z}} &\to \text{Fully connected ($ d_{z}\times 8*8*512  $)}\\      
        &\to \text{256 ConvTran, Stride 1} \to \text{BatchNorm} \to \text{ReLU}\\
        &\to \text{128 ConvTran, Stride 2} \to \text{BatchNorm} \to \text{ReLU}\\
        &\to \text{64 ConvTran, Stride 2} \to \text{BatchNorm} \to \text{ReLU}\\
        &\to \text{3 ConvTran, Stride 2} \to \text{Sigmoid}
    \end{split}
\end{equation*}

\section{Datasets and Training details}
\label{Datasets and Training details}
We list details for each benchmark dataset in following table

\begin{center}
\begin{tabular}{ |l|c|c|c| }
\hline
Datasets&\# Training samples& \# Hold-out samples & Original image size\\
\hline
MNIST&60000&10000&28*28\\
\hline
Fashion-MNIST&60000&10000&28*28\\
\hline
CIFAR-10&50000&10000&32*32\\
\hline
CelebA&162770&19867 &178*218\\
\hline
\end{tabular}
\end{center}

Note that for MNIST,Fashion-MNIST and CIFAR-10, we used default splittings of training sets and testing sets provided in Pytorch (torchvision.datasets). For CelebA, we used default validation set as hold-out samples.

As to the training details, we used same training parameters for all algorithms and datasets, as described in following table

\begin{center}
\begin{tabular}{ |l|c| }
\hline
Latent space dimensions $d_{z}$& 8,16,32,64\\
\hline
Optimizer& Adam with learning rate 3e-4\\
\hline
Batch size& 100\\
\hline
Epochs& 50\\
\hline
\end{tabular}
\end{center}
\textbf{Calculation of Sharpness}

We follow the way in \cite{WAE} in calculating the sharpness of an image. For each generated image, we first transformed it into grayscale and  convolved it with the Laplace filter$\begin{pmatrix}
0 & 1 &0\\
1 & -4 &1\\
0 & 1 &0\\
\end{pmatrix}$, computed the variance of the resulting activations and took the average of all variances. The resulting number is denoted as sharpness (larger is better). The blurrier image will have less edges. As a result, the variance of activations will be small as most activations will be close to zero. Note that we averaged the sharpness of all reconstructed images from hold-out samples for each dataset.

\textbf{Binomial test for two models}

If EVAE and VAE have similar performance in FID score, the probability that EVAE has lower FID score should be $0.5$ in each independent experiment. (Same hypothesis for sharpness). However, according to Table \ref{table mnist} and \ref{table cifar}, EVAE wins 15 experiments for FID score among all datasets. The p-value of winning 15  experiments under null hypothesis is 
\begin{equation*}
    P(X\geq 15)=\binom{16}{16}(0.5)^{16}+\binom{16}{15}(0.5)^{16}\approx 2.6\times 10^{-4} <0.05.
\end{equation*}

P-value is smaller than 0.05 significance level thus EVAE significantly outperforms VAE in FID. Similar calculation can be applied to sharpness, whose p-value of winning  12 experiments is $P(X\geq 15)=\sum_{i=15}^{16}\binom{16}{i}0.5^{16}\approx 2.6\times 10^{-4} <0.05$ and we achieved the same conclusion for the significance of EVAE's superiority in sharpness.





\section{Sampled reconstructed images}
\label{Sampled reconstructed images}

\captionsetup[figure]{font=small}
\begin{figure}[H]
\subfloat[Real images]{\label{control_prob} \includegraphics[width=0.32\columnwidth]{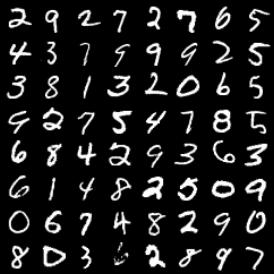}}
    \hfill
    \subfloat[VAE reconstructed images]{\label{small_signal} \includegraphics[width=0.32\columnwidth]{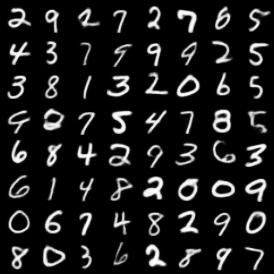}}
    \hfill
\subfloat[EVAE reconstructed images]{\label{mp} \includegraphics[width=0.32\columnwidth]{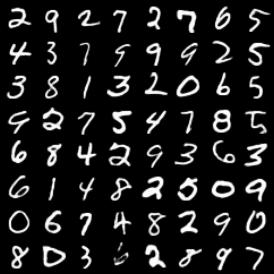}}
\caption{ (a) Sampled real images from hold-out samples in MNIST (b) Reconstructed images by VAE. (c) Reconstructed images by EVAE. Dimension $d_{z}=64$ for both models. See section \ref{Model architecture} for Model architectures.}
\label{mnist sample}
\end{figure}



\captionsetup[figure]{font=small}
\begin{figure}[H]
\subfloat[Real images]{\label{control_prob} \includegraphics[width=0.32\columnwidth]{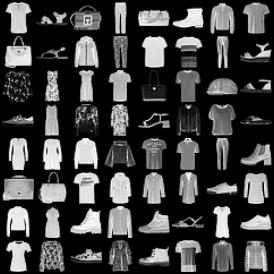}}
    \hfill
    \subfloat[VAE reconstructed images]{\label{small_signal} \includegraphics[width=0.32\columnwidth]{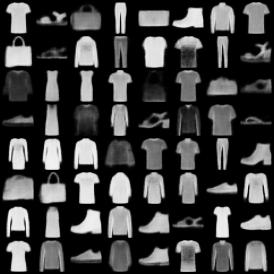}}
    \hfill
\subfloat[EVAE reconstructed images]{\label{mp} \includegraphics[width=0.32\columnwidth]{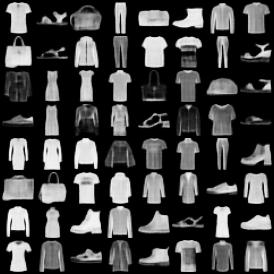}}
\caption{(a) Sampled real images from hold-out samples in Fashion-MNIST (b) Reconstructed images by VAE. (c) Reconstructed images by EVAE. Dimension $d_{z}=64$ for both models.}
\label{fashion-mnist sample}
\end{figure}

\captionsetup[figure]{font=small}
\begin{figure}[H]
\subfloat[Real images]{\label{control_prob} \includegraphics[width=0.32\columnwidth]{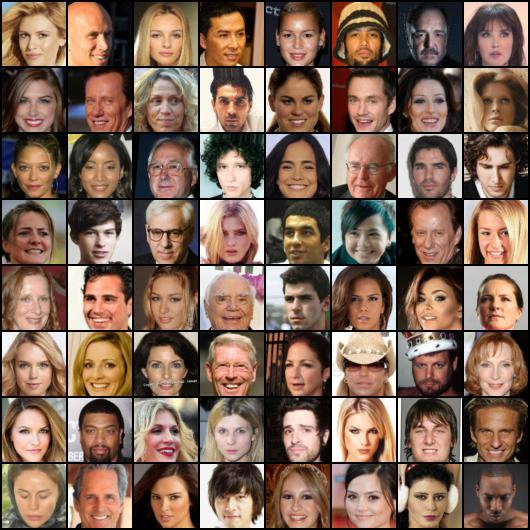}}
    \hfill
    \subfloat[VAE reconstructed images]{\label{small_signal} \includegraphics[width=0.32\columnwidth]{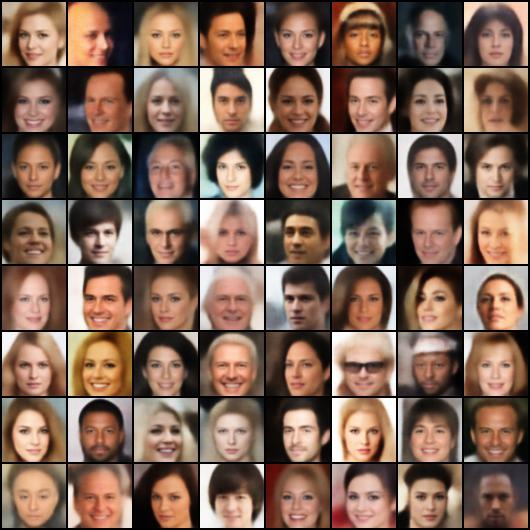}}
    \hfill
\subfloat[EVAE reconstructed images]{\label{mp} \includegraphics[width=0.32\columnwidth]{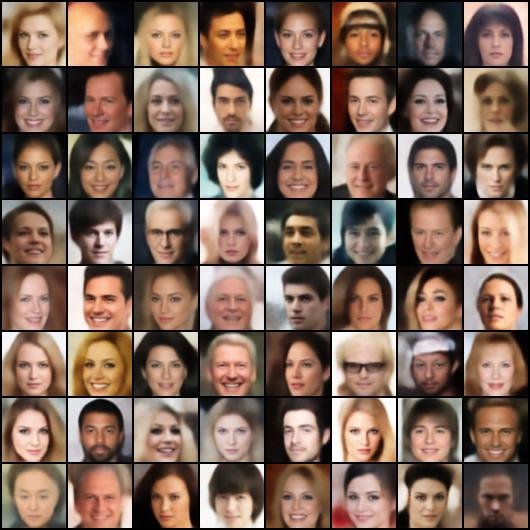}}
\caption{(a) Sampled real images from hold-out samples in CelebA (b) Reconstructed images by VAE. (c) Reconstructed images by EVAE. Dimension $d_{z}=64$ for both models. }
\label{mnist sample}
\end{figure}



\section{Validation curves}
\label{loss curve}

\captionsetup[figure]{font=small}
\begin{figure}[H]
\subfloat[MNIST]{\label{control_prob} \includegraphics[width=0.48\columnwidth]{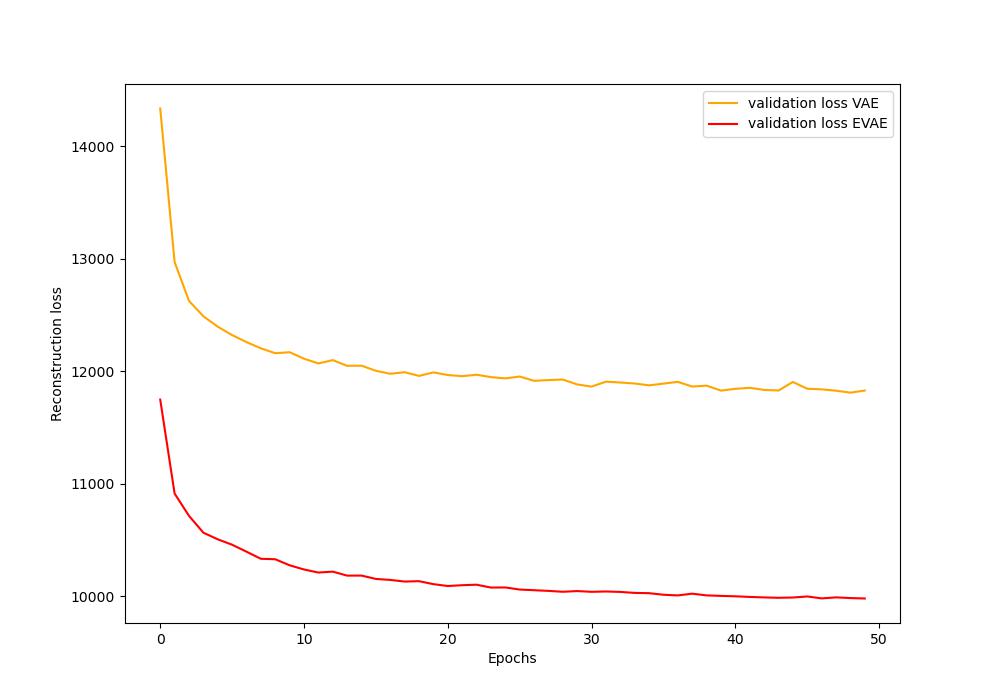}}
    \hfill
    \subfloat[Fashion-MNIST]{\label{small_signal} \includegraphics[width=0.48\columnwidth]{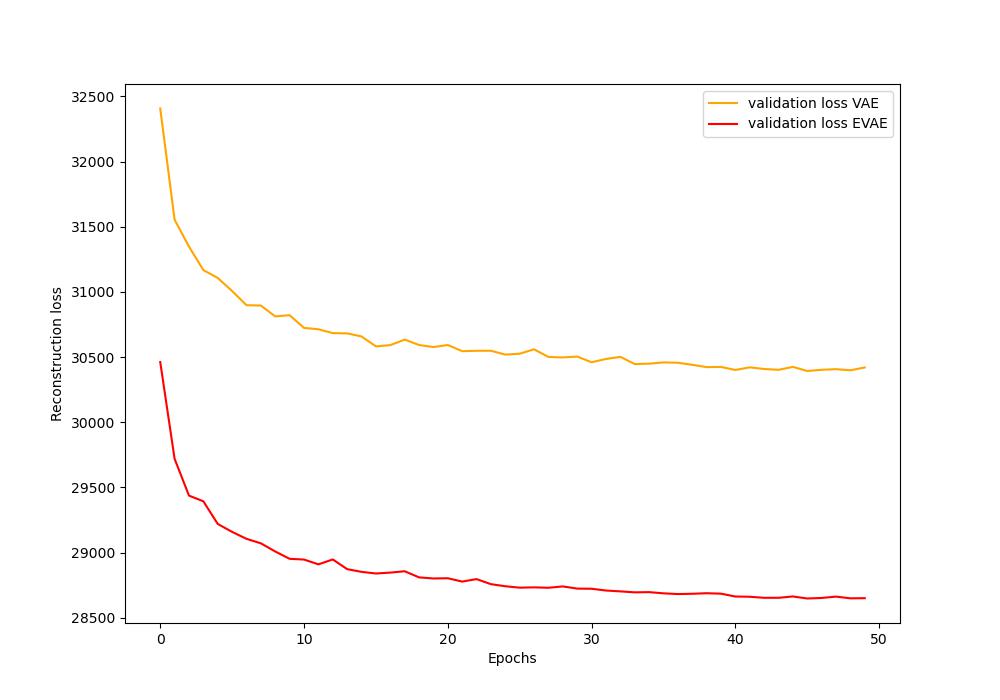}}

\caption{Reconstruction validation loss curves as function of Epochs. ($d_{z}=64)$. Red curve is for EVAE and yellow one represents VAE. Binary cross entropy loss is reduced to sum for each batch (a) MNIST (b) Fashion-MNIST.}
\label{loss curve MNIST Fashion MNIST}
\end{figure}

\captionsetup[figure]{font=small}
\begin{figure}[H]
\subfloat[CIFAR-10]{\label{control_prob} \includegraphics[width=0.48\columnwidth]{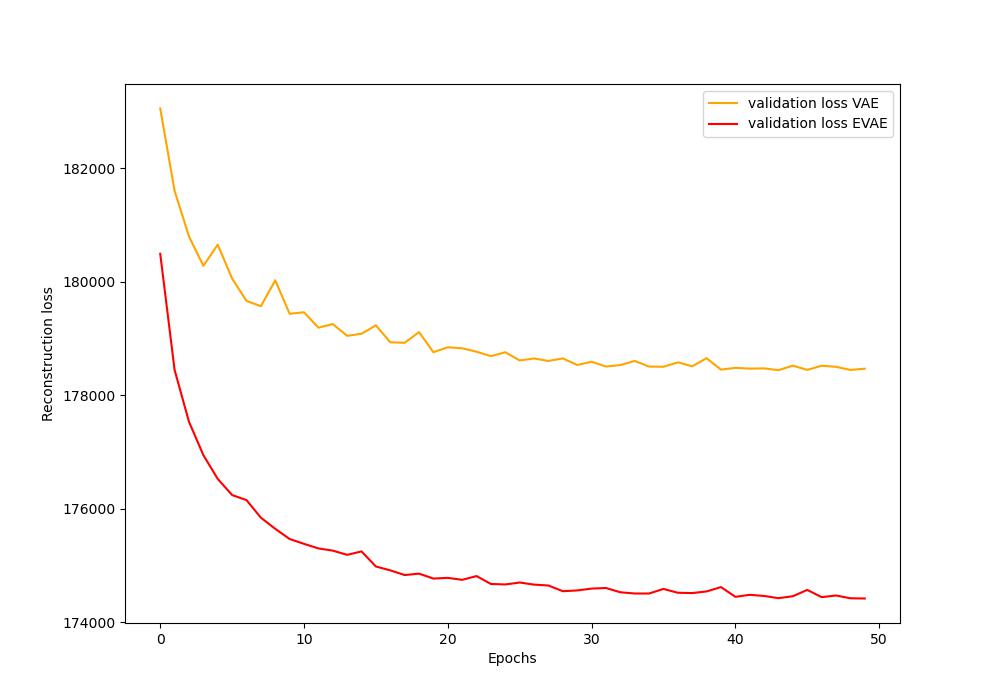}}
    \hfill
    \subfloat[CelebA]{\label{small_signal} \includegraphics[width=0.48\columnwidth]{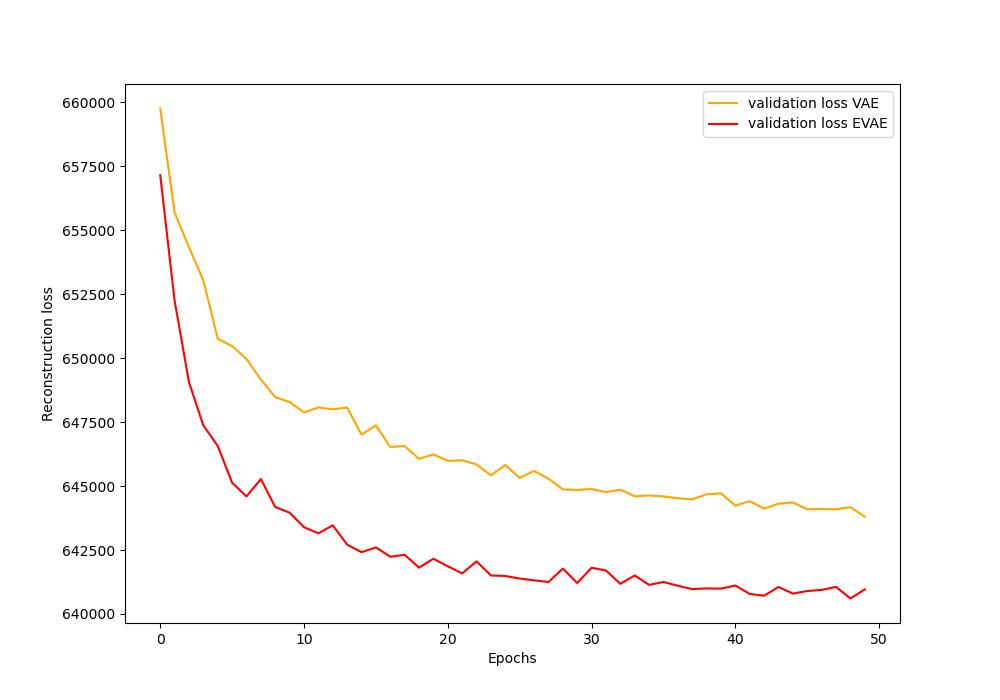}}

\caption{Reconstruction validation loss curves as function of Epochs. ($d_{z}=64)$. Red curve is for EVAE and yellow one represents VAE. Binary cross entropy loss is reduced to sum for each batch (a) CIFAR-10  (b) CelebA. }
\label{loss curve CIFAR CelebA}
\end{figure}

\section{More experiment details on EVAE}
\label{More results}
\subsection{Benchmark datasets}
\label{Benchmark datasets}
 For comparison, we applied a classical CNN architecture (Appendix \ref{Model architecture}) in encoding and decoding part for all datasets. Consequently, the main differences between EVAE and VAE in implementation stem from the resampling step and  target function in training procedure. All FID scores and sharpness are based on hold-out samples.  We also evaluated VAE and EVAE with different dimensions $d_{z}$ of latent spaces.  Table \ref{table cifar} summarize the results  on CIFAR10 adn CelebA datasets with uniform prior support parameter B=0.1 in EVAE. More results for MNIST and Fashion-MNIST can be found in Table \ref{table mnist}

\begin{table}[H]
  \caption{VAE and EVAE results on MNIST and Fashion-MNIST datasets}
  \label{table mnist}
  \centering
 \begin{tabular}{l llll llll llll}
\toprule
 Datasets&\multicolumn{4}{c}{MNIST}
&
\multicolumn{4}{c}{Fashion-MNIST} \\
\cmidrule(l){2-5}\cmidrule(l){6-9}
\multirow{2}{*}{$d_{z}$} &\multicolumn{2}{c}{VAE} &\multicolumn{2}{c}{EVAE}    & \multicolumn{2}{c}{VAE} &\multicolumn{2}{c}{EVAE}      \\
 &  FID  &Sharpness &FID  &Sharpness    & FID  &Sharpness &FID&Sharpness   \\
\midrule

8 &  15.70&0.0211 &14.74& 0.0243&41.47&0.0157 &36.18&0.0176 \\
16 &  11.93&0.0252 &10.49& 0.0305&38.85&0.0159& 26.71&  0.0216 \\
32  & 11.97 &0.0249& 7.92& 0.0338&   38.74& 0.0163&18.47& 0.0251\\
64  & 11.97 &0.0246& 5.13& 0.0360&  39.31&0.0161&12.02& 0.0267\\
\bottomrule
\end{tabular}
\end{table}

\subsection{EVAE with Gaussian prior}
\label{EVAE with gaussian prior}
To check whether the superior performance of EVAE over Gaussian VAE comes from the new prior (uniform) or new posterior (convolution between prior and KDE ), we performed extra experiments for EVAE with Gaussian prior. Table \ref{table mnist g} and Table \ref{table cifar g} in Appendix \ref{EVAE with gaussian prior} compare the FID score and sharpeness of EVAE (with Gaussian prior and B=0.1) with VAE and results show that EVAE still outperforms VAE. To this point, we can say the superior performance of EVAE mainly comes from the introduction of  KDE based posterior.

The experiment details are the same with Appendix \ref{Model architecture}. The only difference is we replaced uniform prior with Gaussian prior and we still kept the support parameter $B (=0.1)$ in front of Gaussian prior.

\begin{table*}[h]
  \caption{VAE and EVAE results on MNIST and Fashion-MNIST datasets}
  \label{table mnist g}
  \centering
 \begin{tabular}{ccccc cccccccc}
\toprule
 Datasets&\multicolumn{4}{c}{MNIST}
&
\multicolumn{4}{c}{Fashion-MNIST} \\
\cmidrule(r){2-5}\cmidrule(l){6-9}
\multirow{2}{*}{$d_{z}$} &\multicolumn{2}{c}{VAE} &\multicolumn{2}{c}{EVAE}    & \multicolumn{2}{c}{VAE} &\multicolumn{2}{c}{EVAE}      \\
 &  FID  &Sharpness &FID  &Sharpness    & FID  &Sharpness &FID&Sharpness   \\
\midrule

8 &  16.14&0.0217 & 15.22& 0.0246&42.22&0.0157 &35.89&0.0181\\
16 &   12.11&0.0248&10.25& 0.0308&38.88&0.0167& 26.65&  0.0218\\
32  & 12.43 &0.0247& 8.22& 0.0335&   39.14&0.0157&18.52& 0.0246\\
64  &11.73 &0.0248& 5.74& 0.0364&   38.96&0.0159&12.18& 0.0271\\
\bottomrule
\end{tabular}
\end{table*}

\begin{table}[H]
  \caption{VAE and EVAE results on CIFAR10 and CelebA datasets}
  \label{table cifar g}
  \centering
 \begin{tabular}{ccccc cccccccc}
\toprule
 Datasets&\multicolumn{4}{c}{CIFAR-10}
&
\multicolumn{4}{c}{CelebA} \\
\cmidrule(r){2-5}\cmidrule(l){6-9}
\multirow{2}{*}{$d_{z}$} &\multicolumn{2}{c}{VAE} &\multicolumn{2}{c}{EVAE}    & \multicolumn{2}{c}{VAE} &\multicolumn{2}{c}{EVAE}      \\
 &  FID  &Sharpness &FID  &Sharpness    & FID  &Sharpness &FID&Sharpness   \\
\midrule

8 & 228.92& 0.0353 &218.84&0.0355&194.36&0.0149 &194.62&0.0145  \\
16 &  183.03& 0.0360& 163.84&0.0355 &144.88&0.0155& 147.78&0.0157     \\
32  & 146.80 &0.0359& 122.17&0.0364&   106.02& 0.0160&102.73& 0.0163\\
64 &  141.22&0.0353& 79.06&0.0356&  80.60&0.0161&65.79& 0.0164\\
\bottomrule
\end{tabular}
\end{table}
\subsection{Effect of uniform prior support parameter B}
\label{effect of B}

Table \ref{effectB} summarizes the performance of EVAE with uniform prior under different values of B in different datasets. ($d=64$ and other settings are the same with original experiments). We can see EVAE becomes worse when B is relatively large. One reason maybe that prior distribution with larger support is likely to generate outlier samples and lower down the quality of reconstructed sample. Additionally, the large value of B will add more weight on the regularization term of  the new target function (\ref{sample ELBO}) of EVAE, which trades off the reconstruction performance of EVAE.

\begin{table}[htbp]
  \caption{The \textbf{FID} score of EVAE with different constant value of B in MNIST, Fashion-MNIST and CIFAR-10 datasets.($d_{z}=64$ in all cases)}
  \label{effectB}
  \centering
 \begin{tabular}{l lll}
\toprule
Support parameter B& MINST & Fashion-MNIST &CIFAR-10\\
\midrule

$\text{B}=0.01$&  4.96& 12.2 &81.1\\
$\text{B}=0.1$& 5.14&12.4 &79.4\\
$\text{B}=1$ & 7.13 &16.9 & 93.9 \\
$\text{B}=10$ &11.29 &33.7& 153.3\\
$\text{B}=20$ &12.93 &43.1& 168.7\\
\bottomrule
\end{tabular}
\end{table}


\subsection{Time efficiency of EVAE}
\label{time efficiency}
One advantage of kernel posterior proposed in equation (\ref{approx posterior}) is that the quadratic functional $I(K)$ has a closed form for many distributions, while the closed form of KL divergence in standard ELBO can be hard to derive when we want to use complicated posterior and prior. For example, the KL divergence becomes piecewise for the uniform prior and posterior. In those cases, time-consuming Monte Carlo simulations might be needed. 

To explore the time efficiency of EVAE, we performed additional experiments in Table \ref{time efficiency} to compare the average and standard error of the training time (in seconds) per epoch (10 epochs total for each experiment) for VAE and EVAE in CIFAR10 and CelebA-64.

\begin{table}[H]
  \caption{Training time of EVAE and VAE per epoch in seconds}
  \label{time efficiency table}
  \centering
 \begin{tabular}{l llllll}

\cmidrule(l){1-5}
\multirow{2}{*}{Latent space dimension $d_{z}$} &\multicolumn{2}{c}{CIFAR10} &\multicolumn{2}{c}{CelebA-64}        \\
 &  VAE(std)  &EVAE(std) &VAE(std)  &EVAE(std)  \\
\midrule

$d_{z}=8$&  7.21(0.28)&7.9(0.19) &208.94(5.95)& 208.42(4.33)\\
$d_{z}=16$& 7.22(0.43)&7.99(0.19) &209.10(5.94)& 210.88(5.35)\\
$d_{z}=32$ & 7.9(0.6) &8.87(0.31)& 208.02(3.22)& 210.23(2.81)\\
$d_{z}=64$ &7.56(0.4) &8.56(0.63)& 206.85(5.24)& 207.72(5.66)\\
\bottomrule
\end{tabular}
\end{table}

In general, we found that EVAE has the comparable time efficiency to VAE. The slight increase in training time for EVAE results from the sampling process of the Epanechnikov kernel, which requires a few more samples to achieve the Epanechnikov density, as described in Algorithm \ref{Sample from EK}. The difference in training time is negligible when the latent space dimension is large, which facilitates the application of EVAE in high-resolution datasets.

\section{Broader Impact Statement}
\label{Broader Impact Statement}
This paper aims to propose a new perspective in modeling the posterior in generative models by kernel density estimations. The theoretical results should not have negative societal impacts. One possible negative impact resulting from EVAE might be the misuse of generative models in producing fake images which may lead to security issues in some face recognition based systems. Few mitigation strategies: (1) gate the release of models for commercial use; (2) add a mechanism for monitoring fake images generated by models such as the discriminator in GAN models. We can also restrict the private datasets used in training the generative model. All benchmark datasets used in this paper are public and well known to the machine learning community.

\end{document}